\documentclass[runningheads]{llncs}
\usepackage{epsf}
\usepackage{graphicx}
\usepackage{comment}
\usepackage{multicol}
\usepackage{amsmath}
\usepackage{multirow}

\usepackage{amssymb}
\usepackage{subfig}
\usepackage{mathtools}
\usepackage[symbol]{footmisc}

\begin{document}

\title{Effectiveness~of~Self Normalizing Neural Networks for Text Classification} 

\author{
Avinash Madasu and Vijjini Anvesh Rao}
\institute{Samsung R\&D Institute, Bangalore\\
           \email{\{m.avinash,a.vijjini\}@samsung.com}}

\maketitle

\begin{abstract}
Self Normalizing Neural Networks(SNN) proposed on Feed Forward Neural Networks(FNN) outperform regular FNN architectures in various machine learning tasks. Particularly in the domain of Computer Vision, the activation function Scaled Exponential Linear Units (SELU) proposed for SNNs, perform better than other non linear activations such as ReLU. The goal of SNN is to produce a normalized output for a normalized input. Established neural network architectures like feed forward networks and Convolutional Neural Networks(CNN) lack the intrinsic nature of normalizing outputs. Hence, requiring additional layers such as  Batch Normalization. Despite the success of SNNs, their characteristic features on other network architectures like CNN haven't been explored, especially in the domain of Natural Language Processing. In this paper we aim to show the effectiveness of proposed, Self Normalizing Convolutional Neural Networks(SCNN) on text classification. We analyze their performance with the standard CNN architecture used on several text classification datasets. Our experiments demonstrate that SCNN achieves comparable results to standard CNN model with significantly fewer parameters. Furthermore it also outperforms CNN with equal number of parameters.

\end{abstract}
\section{Introduction}
The aim of Natural Language Processing(NLP) is to analyze and extract information from textual data in order to make computers understand language, the way humans do. Unlike images which lack sequential patterns, texts involve amplitude of such information which makes processing very distinctive.
\newline
The level of processing varies from paragraph level, sentence level, word level and to the character level. Deep neural network architectures achieved state-of-art results in many areas like Speech Recognition \cite{hinton2012deep} and Computer Vison \cite{krizhevsky2012imagenet}. The use of neural networks in Natural language processing can be traced back to \cite{rumelhart1986learning} where the backpropogation algorithm was used to make networks learn familial relations. The major advancement was when \cite{bengio2003neural} applied neural networks to represent words in a distributed compositional manner. \cite{mikolov2013efficient} proposed two neural network models CBoW and Skip-gram for an efficient distributed representation of words. This was a major break-through in the field of NLP. From then, neural network architectures achieved state-of-results in many NLP applications like Machine Translation \cite{bahdanau2014neural}, Text Summarization \cite{rush2015neural} and Conversation Models \cite{vinyals2015neural} .

Convolutional Neural Networks \cite{lecun1998gradient} were devised primarily for dealing with images and have shown remarkable results in the field of computer vision\cite{articleimagenet,simonyan2014very}. In addition to their contribution in Image processing, their effectiveness in Natural language processing has also been explored and shown to have strong performance in Sentence\cite{kim2014convolutional} and Text Classification\cite{conneau2016very}.

The intuition behind Self Normalizing Neural Networks(SNN) is to drive neuron activations across all layers to emit a zero mean and unit variance output. This is done with the help of the proposed activation in SNNs, SELU or scaled exponential linear units. With the help of SELUs an effect alike to batch normalization is replicated, hence slashing the number of parameters along with a robust learning. Special Dropouts and Initialization also help in this learning, which make SNNs remarkable to traditional Neural Networks. As Image based inputs and Text based inputs differ from each other in form and characteristics, in this paper we propose certain revisions to the SNN architecture to empower them on texts efficiently. 

In this paper, to explore effectiveness of self normalizing neural networks in text classification, we propose an architecture, Self Normalizing Convolutional Neural Network (SCNN) built upon convolutional neural networks. A thorough study of SCNNs on various benchmark text datasets, is paramount to ascertain importance of SNNs in Natural Language Processing.

\section{Related Work}
\label{sec:relatedwork}

Prior to the success of deep learning, text classification heavily relied on good feature engineering and various machine learning algorithms.

Convolutional Neural Networks \cite{lecun1998gradient} were devised primarily for dealing with images and have shown remarkable results in the field of computer vision\cite{articleimagenet,simonyan2014very}. In addition to their contribution in Image processing, their effectiveness in Natural language processing has also been explored and shown to have strong performance. Kim \cite{kim2014convolutional} represented an input sentence using word embeddings that are stacked into a two dimensional structure where length corresponds to embedding size and height with average sentence length. Processing this structure using kernel filters of fixed window size and max pooling layer upon it to capture the most important information has shown them promising results on text classification. Additionally, very deep CNN architectures \cite{conneau2016very} have shown state-of-the art results in text classification, significantly reducing the error percentage. As CNNs are limited to fixed window sizes, \cite{rcnn2015} have proposed a recurrent convolution architecture to exploit the advantages of recurrent structures that capture distant contextual information in ways fixed windows may not be able to.

Klambauer \cite{NIPS2017_6698} proposed Self Normalizing Neural Networks (SNN) upon feed forward neural networks, significantly outperformed FNN architectures on various machine learning tasks. Since then, the activation proposed in SNNs, SELU have been widely studied in Image Processing\cite{lguensat2018eddynet,zhang2017deformable,goh2017smiles2vec}, where they have been applied on CNNs to achieve better results. SELU's effectiveness have also been explored in Text\cite{kumar2017sentiment,rosa2017retuyt,meisheri2017sentiment} processing tasks. However these applications are limited to applying just SELUs in their \cite{lguensat2018eddynet,zhang2017deformable,goh2017smiles2vec,kumar2017sentiment,rosa2017retuyt,meisheri2017sentiment} respective architectures.

\section{Self-Normalizing Neural Networks}
Self-Normalizing Neural Networks(SNN) are introduced by Günter Klambauer \cite{NIPS2017_6698} to learn higher level abstractions. Regular neural network architectures like Feed forward Neural Networks(FNN), Convolutional Neural Networks(CNN) lack the property of normalizing outputs and require additional layers like Batch Normalization\cite{43442} for normalizing hidden layer outputs. SNN are specialized neural networks in which the neuron activations automatically converge to a fixed mean and variance. Training of deep CNNs can be efficiently stabilized by using batch normalization and by using Dropouts \cite{JMLR:v15:srivastava14a}. However FNN suffer from high variance when trained with these normalization techniques. In contrast, SNN are very robust to high variance thereby inducing variance stabilization and overcoming problems like exploding gradients\cite{NIPS2017_6698}.
SNN differs from naive FNN by the following:
\subsection{Input Normalization}
To get a normalized output in SNN without requiring layers like batch normalization, the inputs are normalized.
\subsection{Intialization}
Weights initialization is an important step in training neural networks.Several initialization methods like glorot uniform\cite{pmlr-v9-glorot10a} and lecun normal\cite{LeCun:1998:EB:645754.668382} have been proposed. FNN and CNN are generally initialized using glorot uniform whereas SNN are initialized using lecun normal. Glorot uniform initialization draws samples centered around 0 and with standard deviation as:
\begin{equation}
    stddev = \sqrt{\frac{2}{(in+out)}}
\end{equation}
Lecun normal initialization draws samples centered around 0 and with standard deviation as:
\begin{equation}
    stddev = \sqrt{\frac{1}{in}}
\end{equation}
where $in$ and $out$ represent dimensions of weight matrix corresponding to number of nodes in previous and current layer respectively.
\subsection{SELU activations}
Scaled exponential linear units (SELU) is the activation function proposed in SNNs. In general, FNN and CNN use rectified linear units(ReLU) as activation. ReLU activation clips negative values to 0 and hence suffers from dying ReLU problem\footnote{ http:
//cs231n.github.io/neural-networks-1/}.  As explained by \cite{NIPS2017_6698} an activation function should contain both positive and negative values for controlling mean, saturation regions for reducing high variance and slope greater than one to increase variance if its value is too small. Hence SELU activation is introduced to preserve the aforementioned properties. SELU activation function is defined as :
\begin{equation}
   selu(x) = \lambda {\begin{cases} 
        x & \text{if}\ x > 0\\ 
        \alpha e^{x}-\alpha & \text{if}\ x \leqslant 0 \end{cases}}
\end{equation}
    where $x$ denotes input, $\alpha$ ($\alpha = 1.6733$), $\lambda$ ($\lambda$ = 1.0507) are hyper parameters, and e stands for exponent
\subsection{Alpha Dropout}
\quad Standard dropout\cite{JMLR:v15:srivastava14a} drops neurons randomly by setting their weights to 0  with a probability $1-p$ . This prevents the network to set mean and variance to a desired value. Standard dropout works very well with ReLUs because in them, zero falls under the low variance region and is the default value. For SELU, standard dropout does not fit well because the default low variance is $\lim_{x \to \infty} selu(x)$ = - $\lambda \alpha$ = $\alpha^{'}$\cite{NIPS2017_6698}.
Hence alpha dropout is proposed which sets input values randomly to $\alpha^{'}$. Alpha dropout restores the original values of mean and variance thereby preserving the self-normalizing property \cite{NIPS2017_6698}. Hence, alpha dropout suits SELU by making activations into negative saturation values at random.

\section{Model}
We propose Self-normalizing Convolutional Neural Network (SCNN) for text classification as shown in the figure \ref{fig:scnn}. 
To show the effectiveness of our proposed model, we adapted the standard CNN architecture used for text classification to SCNN with the following changes:
\subsection{Word Embeddings are not normalized}
\label{sec:notnorm}
Self-Normalizing Neural Networks require inputs to be normalized for the outputs to be normalized\cite{NIPS2017_6698}. Normalization of inputs work very well in computer vision because images are represented as pixel values which are independent of the neighbourhood pixels. In contrast, word embedding for a particular word is created based on its co-occurrence with context. Words exhibit strong dependency with their neighbourhood words. Similar words contain similar context and are hence close to each other in their embedding space. If word embeddings are normalized, the dependencies are disturbed and their normalized values will not represent the semantics behind the word correctly.
\subsection{ELU activation as an alternative to SELU}

\begin{figure}%
\label{fig:stats}
    \centering
    \subfloat[Accuracy Score comparison on MR, SO, IMDB, TREC Datasets ]{{\includegraphics[width=150px]{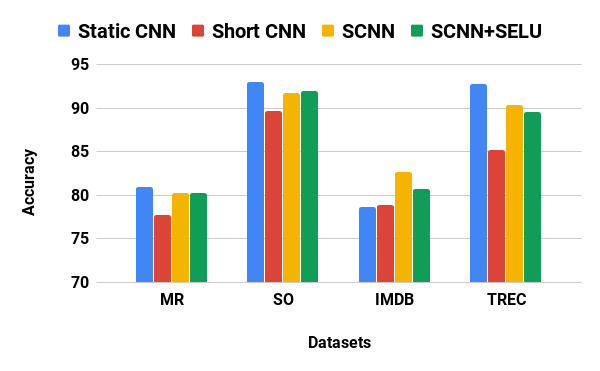} }}%
    \qquad
    \subfloat[F1-Score comparison on CR, MPQA Datasets]{{\includegraphics[width=150px]{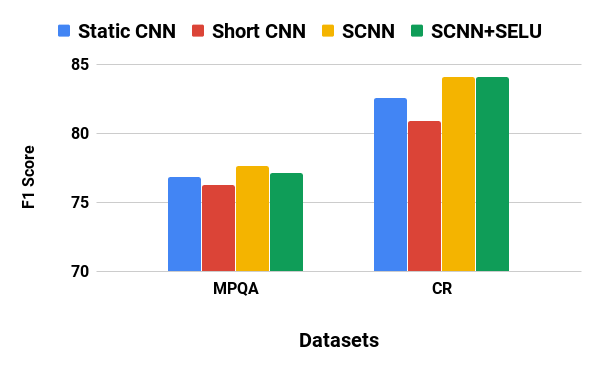} }}%
  \qquad
    \subfloat[Performance comparison of SELU and ELU on MR, SO, IMDB, TREC Datasets]{{\includegraphics[width=150px]{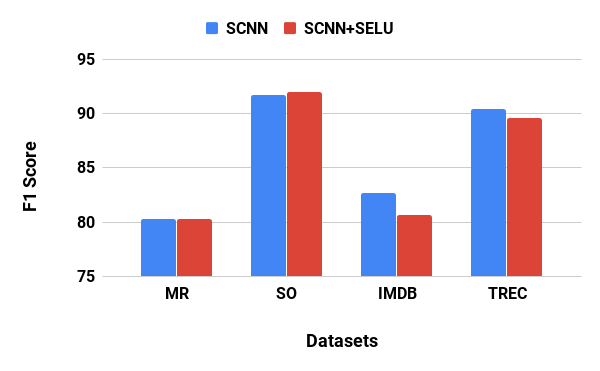} }}%
    \qquad
    \subfloat[Performance comparison of SELU and ELU on CR, MPQA Datasets Length]{{\includegraphics[width=150px]{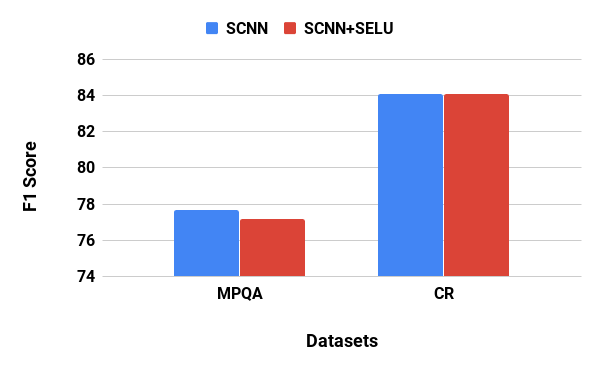}}}%
    \caption{Figures to demonstrate performance of all models}%
    \label{fig:example}%
\end{figure}

SELU activation originally proposed for SNN\cite{NIPS2017_6698} preserve the properties of SNN if the inputs are normalized. When inputs are normalized, applying SELU on the activations does not shift the mean. However if inputs weren't normalized, due to the parameter $\lambda$ in SELU activation, the neuron outputs will be scaled by a factor $\lambda$ thereby shifting the mean and variance to a value away from the desired mean and variance. These values are further propagated to other layers thereby shifting mean and variance more and more. Since input word embeddings cannot be normalized as explained in section \ref{sec:notnorm}, we use ELU activation \cite{DBLP:journals/corr/ClevertUH15} in the proposed SCNN model instead. ELU activation function is defined as :
\begin{equation}
   elu(x) = {\begin{cases} 
        x & \text{if}\ x > 0\\ 
        \alpha e^{x}-\alpha & \text{if}\ x \leqslant 0 \end{cases}}
\end{equation}
    where $\alpha$ is a hyper parameter, and e stands for exponent. The absence of parameter $\lambda$ in ELU prevents greater scaling of neuron outputs. ELU activation pushes the mean of the activations closer to zero even if inputs are not normalized which enable faster learning \cite{DBLP:journals/corr/ClevertUH15}. We compare the performance of SCNN with both SELU and ELU activations and the results presented in table \ref{table:results} and figure \ref{fig:example}.
\subsection{Model Architecture}

\begin{figure}
\begin{center}
\includegraphics[width=\textwidth]{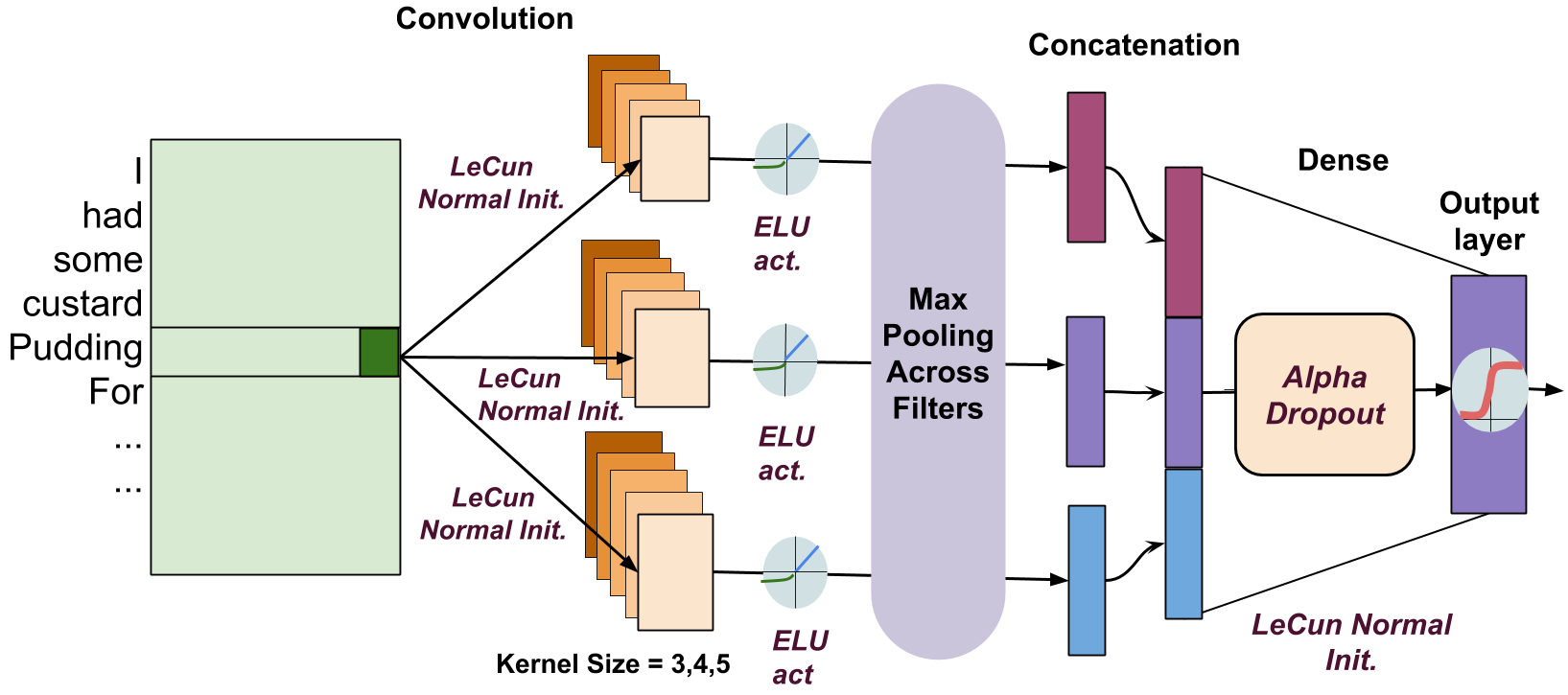}
\caption{Architecture of proposed SCNN model} \label{fig:scnn}
\end{center}
\end{figure}

The SCNN architecture is shown in the figure \ref{fig:scnn}.
Let V be the vocabulary size considered for each dataset and $X$ $\in$ $\mathbb{R}^{V \times d}$ represent the word embedding matrix where each word $X_{i}$ is a $d$ dimensional word vector. Words present in the pretrained word embedding\footnote{https://code.google.com/archive/p/word2vec/} are assigned their corresponding word vectors. Word that are not present are initialized to 0s. Based on our experiments, SCNN showed better performance when absent words are initialized to 0s than randomly initialization. A maximum sentence length of $N$ is considered per sentence or paragraph. If the sentence or paragraph length is less than $N$, zero padding is done. Therefore, $I$ $\in$ $\mathbb{R}^{N \times d}$ dimensional vector per each sentence or paragraph is provided as input to the SCNN model.

Convolution operation is applied on $I$ with kernel $K$ $\in$ $\mathbb{R}^{h \times d} $(h $\in$  \{3,4,5\}) is applied to input vectors with a window size of $h$. The weight initialization of these kernels is done using lecun normal \cite{LeCun:1998:EB:645754.668382} and bias is initialized to 0. A new feature vector is $C$ $\in $ $\mathbb{R}^{(N-h+1) \times 1 }$ is obtained after the convolution operation for each filter.
\begin{equation}
    C = f(I \circledast K)
\end{equation}
where $f$ represents the activation function (ELU).
Number of convolution filters vary depending on the dataset, table \ref{tab:parameters} summarizes the number of parameters for all of our experiments.
Maxpooling operation is applied across each filter $C$ to get the maximum value. The outputs from the maxpooling layer across all filters are concatenated. Alpha dropout \cite{NIPS2017_6698} with a dropout value 0.5 is applied on the concatenated layer. The concatenated layer is densely connected to the output layer with activation as sigmoid if task is binary classification and softmax otherwise. 
\section{Experiments and Datasets}
\begin{table}
\setlength{\tabcolsep}{8pt}
\caption{Summary statistics of all datasets}
\label{tab:statistics}
\begin{center}
\begin{tabular}{c c c c} 
\hline
Datasets & No. of classes&Dataset Size & Test \\
\hline
MR & 2&10662 & 10 fold Cross Validation\\
SO & 2&10000 & 10 fold Cross Validation\\
IMDB &2&50000 & 25000\\
TREC &6&5952 & 500\\
CR &2&3773 & 10 fold Cross Validation\\
MPQA &2&10604 & 10 fold Cross Validation\\
\hline
\end{tabular}
\end{center}
\end{table}
\begin{table}
\setlength{\tabcolsep}{8pt}
\caption{Parameters of examined models on all datasets}
\label{tab:parameters}
\begin{center}
\begin{tabular}{c c c c c} 
\hline
\multirow{3}{*}{\textbf{Datasets}} &  \multicolumn{2}{c}{\textbf{No of Conv.Filters}} & \multicolumn{2}{c}{\textbf{No of Parameters}}\\
   & SCNN and & \multirow{2}{*}{Static CNN\cite{kim2014convolutional}} &  SCNN and & \multirow{2}{*}{Static CNN\cite{kim2014convolutional}} \bf \\
      & Short-CNN & & Short-CNN & \bf \\
\hline
\hline
MR  & 210 &300 & $\approx 254k$& $\approx 362k$ \\
SO &  210 &300& $\approx 254k$& $\approx 362k$\\
IMDB  & 210  &300 & $\approx 254k$ & $\approx 362k$\\
TREC &  210 &300 &  $\approx 254k$ & $\approx 362k$\\
CR  & 90 &300  &$\approx 108k$& $\approx 362k$ \\
MPQA & 90 &300 &$\approx 108k$ & $\approx 362k$\\
\hline 
\end{tabular}
\end{center}
\end{table}
\subsection{Datasets}
We performed experiments on various benchmark data sets of text classification. The summary statistics for the datasets are shown in table \ref{tab:statistics}
\subsubsection{Movie Reviews(MR)}
It consists of 10662 movie reviews with 5331 positive and 5331 negative reviews\cite{Pang+Lee:05a}. Task involves classifying reviews into positive or negative sentiment.
\subsubsection{Subjectivity Objectivity(SO)}
The dataset consists of 10000 sentences with 5000 subjective sentences and 5000 objective \cite{Pang+Lee:04a}. It is a binary classification task of classifying sentences as subjective or objective.
\subsubsection{IMDB Movie Reviews(IMDB)}
The dataset consists of 50000 movie reviews of which 25000 are positive and 25000 are negative \cite{maas-EtAl:2011:ACL-HLT2011}.  
\subsubsection{TREC}
TREC dataset contains questions of 6 categories based on the type of question: 95 questions for Abbreviation,1344 questions for Entity, 1300 questions for 
Description, 1288 questions for Human, 916 questions for Location and 1009 questions for Numeric Value \cite{Li:2002:LQC:1072228.1072378}.
\subsubsection{Customer Reviews(CR)}
The dataset consists of 2406 positive reviews and 1367 negative reviews. It is a binary classification task of predicting positive or negative sentiment \cite{hu2004mining}.
\subsubsection{MPAQ}
The dataset consists of 3311 positive reviews and 7293 negative reviews. Binary classification task of predicting positive and negative opinion \cite{wiebe2005annotating}.
\subsection{Baseline Models}
We compare our proposed model SCNN with the following models:
\subsubsection{Static CNN model}
We compare SCNN with the static model, one of the standard CNN models proposed for text classification \cite{kim2014convolutional}.
\subsubsection{SCNN with SELU activation}
SNN was originally proposed with SELU activation. We performed experiments on SCNN using SELU as the activation function in place of ELU.
\subsubsection{Short CNN}
Our model SCNN is proposed with fewer parameters compared to Static CNN model\cite{kim2014convolutional}. To show the effectiveness of SCNN, we perform experiments on Static CNN model with same number of parameters as SCNN, we refer this model as Short CNN.

\subsection{Model parameters}
Table \ref{tab:parameters} shows the parameter statistics for all the models. SCNN and Short CNN models are experimented using same number of parameters. We used 70 convolution filters for each kernel in case of MR, SO, IMDB and TREC datasets. For datasets CR and MPQA we considered 30 filters for each kernel. In MPQA dataset, the average sentence length is 3. Hence, we reduced the convolution filters from 70 to 30.
\subsection{Training}
We process the dataset as follows: Each sentence or paragraph is converted to lower case. Stop words are not removed from the sentences. We consider a vocabulary size $V$ for each dataset based on the word counts. The datasets IMDB, TREC have predefined test data and for other datasets we used 10-fold Cross Validation.
The parameters chosen vary depending on the dataset size. 
Table \ref{tab:parameters} shows the parameters of SCNN model for all datasets.
We used Adam \cite{DBLP:journals/corr/KingmaB14} as the optimizer for training SCNN.
\section{Results and Discussion}

\subsection{Results}
The performance of the models for all datasets is shown in table \ref{table:results}. For all the balanced datasets MR, SO, IMDB and TREC, accuracy is used as the metric for comparison. The performance comparison on imbalanced datasets like CR and MPQA cannot be justified using accuracy because imbalance can induce bias in the models' prediction. Hence we use F1-Score as the metric for performance analysis for CR and MPQA. The datasets IMDB and TREC have preexisting train, test sets. Therefore, we report our results on the provided test sets for them. For remaining datasets, we report results using 10-fold cross validation(CV).
\subsection{Discussion}
\subsubsection{SCNN models against Short-CNN:}\hfill \newline
 When we compare SCNN with SELU and SCNN to Short CNN, both the models of SCNN outperform Short CNN for all the datasets. This shows that SCNN models perform better than CNN models (Short CNN) with same number of parameters indicating a better generalization of training. There is a significant improvement in accuracy and F1-Score when SCNN models are used in place of CNN. We believe that the use of activation functions ELU and SELU in the SCNN models as opposed to ReLU  is the leading factor behind  this performance difference between SCNN and CNN. In particular, ReLU activation suffers from dying ReLU problem\footnote{ http:
//cs231n.github.io/neural-networks-1/}. In ReLU, the negative values are cancelled to 0. Therefore negative values in the pretrained word vectors are ignored  thereby loosing information about negative values. This problem is solved in ELU and SELU by having activation even for the negative values. In comparison to ReLU, ELU and SELU have faster and accurate training convergence leading to better generalization performance.

\subsubsection{ELU against SELU in SCNN:}\hfill \newline
We proposed SCNN using ELU as activation function opposed to SELU, the activation function introduced originally for SNN. We found that if ELU is used as activation, the performance of SCNN is better for a majority of the datasets. Our results from table \ref{table:results} substantiate the claim about the effectiveness of ELU activation. The characteristic
 difference between SELU and ELU activations is the parameter $\lambda$ ($\lambda > 1$) in SELU activation which scales the neuron outputs. SELU is effective for maintaining normalized mean and variance when the inputs are normalized. Since, pretrained word vectors are not normalized, the parameter $\lambda$ adversely scales the outputs. This results in a shifted mean and variance from the desired values. On propagation through subsequent layers the difference only gets further magnified. On the other hand, ELU pushes the activations to unit mean even if the inputs are not normalized. Hence, ELU achieved better results compared to SELU activation in SCNN.

\subsubsection{SCNN against static CNN:}\hfill \newline
Our results from table \ref{table:results} indicate that SCNN achieves comparable results to Static CNN. As shown in table \ref{tab:parameters}, the stark difference in the parameter counts between SCNN and static CNN  is more than a million. For datasets IMDB, CR and MPQA, SCNN outperforms static CNN. In case of MR dataset, performance difference between SCNN and static CNN is very minimal.

\begin{table}[!t]
\setlength{\tabcolsep}{8pt}
\caption{Performance of the models on different datasets}
\label{table:results}
\begin{center}
\begin{tabular}{c c c c c c c}
\hline
\multirow{3}{*}{\textbf{Model}} & \multicolumn{6}{c}{\textbf{Datasets}}\\
& MR  & SO & IMDB & TREC & CR & MPQA\\
\hline
\\[-1em]
 &\multicolumn{4}{c}{Accuracy} & \multicolumn{2}{c}{F1-Score}\\
\hline
\hline
\\[-1em]
Short CNN & 77.762 & 89.63 &78.84 &85.2 &76.246 &80.906\\
SCNN w/SELU & 80.266& \textbf{91.99}&80.664 &89.6 &77.166 &84.062\\
SCNN & \textbf{80.308}& 91.759 & \textbf{82.708} & \textbf{90.4}&\textbf{77.666} &\textbf{84.068}\\
\hline
\\[-1em]
CNN-static\cite{kim2014convolutional} & 81 & 93 &78.692 &92.8 &76.852 &82.584\\

\hline
\end{tabular}
\end{center}
\end{table}

\section{Conclusion}
We propose SCNN for performing text classification. Our observations indicate that  SCNN has comparable performance to CNN (Static-CNN \cite{kim2014convolutional}) model with substantially lesser parameters. Moreover SCNN performs significantly better than CNN with equal number of parameters. The experimental results demonstrate the effectiveness of self normalizing neural networks in text classification. Currently, SCNN is proposed with relatively simple architectures. Our work can be further extended by experimenting SCNN on deep architectures. In addition to this, SNN can also be applied on recurrent neural networks(RNN) and its performance can be analyzed.

\bibliographystyle{splncs}
\bibliography{paper}

\end{document}